\documentclass[final,onefignum,onetabnum]{siamart190516}

\usepackage{graphicx}
\usepackage{float}
\usepackage{amsfonts}
\usepackage{amsmath}
\usepackage{amssymb}
\usepackage{bm}
\usepackage{color}
\usepackage[T1]{fontenc}
\usepackage[utf8]{inputenc}
\usepackage{comment}
\usepackage{mathtools}
\usepackage{xcolor}
\usepackage{bbm}
\usepackage{multirow}
\usepackage{subcaption}

\usepackage{lipsum}
\usepackage{amsfonts}
\usepackage{graphicx}
\usepackage{epstopdf}
\usepackage{algorithmic}

\newcommand{\R}{\mathbb{R}} 
\newcommand{\N}{\mathbb{N}}
\newcommand{\bs}{\boldsymbol}

\newsiamremark{remark}{Remark}
\newsiamremark{hypothesis}{Hypothesis}
\crefname{hypothesis}{Hypothesis}{Hypotheses}
\newsiamthm{claim}{Claim}

\DeclareMathOperator*{\argmin}{argmin}

\ifpdf
\hypersetup{
  pdftitle={Score-Oriented Loss (SOL) functions},
  pdfauthor={Francesco Marchetti, Sabrina Guastavino, Michele Piana, and Cristina Campi}
}
\fi

\begin{document}

\title{Score-oriented loss (SOL) functions}

\author{
Francesco Marchetti\thanks{Dipartimento di Matematica \lq\lq Tullio Levi-Civita\rq\rq, Università di Padova, Italy} \and
Sabrina Guastavino\thanks{MIDA, Dipartimento di Matematica, Universit\`a di Genova, Genova, Italy} \and
Michele Piana\thanks{MIDA, Dipartimento di Matematica, Università di Genova and CNR - SPIN Genova, Genova, Italy} 
\and
Cristina Campi\footnotemark[2]
}

\maketitle

\begin{abstract}
Loss functions engineering and the assessment of forecasting performances are two crucial and intertwined aspects of supervised machine learning. This paper focuses on binary classification to introduce a class of loss functions that are defined on probabilistic confusion matrices and that allow an automatic and {\em{a priori}} maximization of the skill scores. The performances of these loss functions are validated during the training phase of two experimental forecasting problems, thus showing that the probability distribution function associated with the confusion matrices significantly impacts the outcome of the score maximization process.
\end{abstract}

\begin{keywords}
  supervised machine learning, binary classification, loss functions, skill scores
\end{keywords}

\begin{AMS}
  68T05, 68Q32, 92B20, 65K10 
\end{AMS}

\section{Introduction}\label{sec:intro}

Neural networks (NNs) are well-established models in the field of machine learning. Given their flexibility and capability in addressing complex tasks where a large amount of data is at the disposal, \textit{deep} NNs are the state-of-art of many classification issues, including image recognition \cite{Pelletier19}, speech analysis \cite{Deng13} and medical diagnostics \cite{kermany2018identifying}; for a detailed treatment of various NNs architectures, as well as of related deep learning topics, we refer e.g. to \cite{Goodfellow16}.

In general, a network depends on a set of weights that are optimized in such a way that an objective function, made of a loss function and a regularization term, is minimized \cite{girosi1995regularization}. Loss functions measure the possible discrepancy between the NN prediction and a label encoding the information about the event occurrence. Loss design is a crucial aspect of machine learning theory \cite{rosasco2004loss} and, over the years, several functions have been proposed, typically with the aim of accounting for the specific properties of the problem under analysis \cite{Janocha16,Lin17,Liu16,Zhu20}. Among these choices, Cross Entropy (CE) and its generalizations \cite{Good52,Lu19,Zhang18} possess a solid theoretical background that can be traced back to information theory, they being related to the notions of \textit{entropy} and \textit{Kullback-Leibler divergence}, and their minimization corresponds to the maximization of the likelihood under a Bernoulli model (see e.g. \cite[Section 2.8]{Murphy12}). 

While the network is optimized by minimizing a certain loss, the classification results are then evaluated by considering some well-established \textit{metrics} or \textit{skill scores}, which are often chosen according to the particular task and derived from the elements of the so-called \textit{confusion matrix} (CM). Although loss minimization and score maximization are two intertwined concepts, yet a direct maximization of a score during the training of the network is to be avoided, since the score is typically a discontinuous function with respect to the predictions given by the model. Therefore, other strategies have been taken into consideration.

For example, in \cite{Huang19} the authors propose an adaptive loss alignment with respect to the evaluation metric via a reinforcement learning approach. A loss function that \textit{approximates} the \textit{0-1 loss} is proposed in \cite{Singh10}. In \cite{Rezatofighi19}, a loss that generalizes the \textit{Jaccard index} is discussed for object detection issues. In order to address the lack of regularity of the score, \textit{fuzzy} or \textit{probabilistic} confusion matrices have been examined e.g. in \cite{Koco13}, where the minimization of the off-diagonal elements of a probabilistic confusion matrix is considered; in \cite{Wang13}, where a confusion entropy metric is constructed; and in \cite{Trajdos16}, where local fuzzy CMs have been employed.

This fuzzy/probabilistic approach was inspirational for the framework introduced in the present paper, which also presents similarities with the main idea investigated in \cite{Ohsaki17} in the context of kernel logistic regression, and which has to be intended as the theoretically-founded generalization of the approach carried out, e.g., in \cite{Zhang17}. Indeed, our study addresses the problem of filling the gap between score maximization and loss minimization by constructing a class of Score-Oriented Loss (SOL) functions out of a classical CM. The main idea consists in treating the threshold that influences the entries of the confusion matrix not as a fixed value, but as a random variable. Therefore, by considering the expected value of the matrix, we build losses that are indeed derivable with respect to the weights of the model. Furthermore, we theoretically motivate and then numerically show that the probability distribution chosen \textit{a priori} for the threshold severely influences the outcome of the \textit{a posteriori} score maximization. More in details, the optimal threshold is very likely to be placed in the \textit{concentration areas} of the \textit{a priori} distribution. As a consequence, using a SOL function related to a certain skill score, we automatically obtain an optimized value without performing any further \textit{a posteriori} procedure.

The paper is organized as follows. In Section \ref{sec:costruzione_loss}, starting from a classical confusion matrix, we construct the theoretical foundation of our SOL functions and we investigate their properties. Section \ref{sec:results} describes two possible choices for the probability distribution associated to the threshold and illustrates their application to two experimental forecasting problems. Finally, our conclusions are offered in Section \ref{sec:conclusions}.

\section{Construction and properties of score-oriented losses}\label{sec:costruzione_loss}

Let $\Omega\subset\mathbb{R}^m$ and let $\mathcal{X}\subset \Omega$ be a set of data, $m\in\mathbb{N}_{\ge 1}$. Let $\mathcal{Y}$ be a finite set of $d$ \textit{labels} or \textit{classes} to be learned, where $d\in\mathbb{N}_{\ge 2}$ and the labels (or classes) are integer encoded or \textit{one-hot} encoded in practice \cite{Harris13}. Suppose that each element in $\mathcal{X}$ is uniquely assigned to a class in $\mathcal{Y}$. Then, the supervised classification problem, where $d=2$, consists of constructing a function $g$ on $\Omega$ by \textit{learning} from the labeled data set $\mathcal{X}$, in such a way that $g$ \textit{models} the data-label relation between elements in $\Omega$ and labels in $\mathcal{Y}$. In order to achieve such result, one usually defines a \textit{loss} function $\ell(g(\boldsymbol{x}),y)$ that measures the possible discrepancy between the prediction $g(\boldsymbol{x})$ given by the model, $\bs{x}\in\mathcal{X}$, and its true label $y\in\mathcal{Y}$. In this view, an NN builds the function $g$ by minimizing the loss on the training set $X$, as well as focusing on its generalization capability when predicting possible unseen test samples in $\Omega$.

More specifically, the network produces the output
\begin{equation*}
    \hat{y}(\bs{x},\bs{w})\coloneqq (\sigma\circ  h)(\bs{x},\bs{w})\in [0,1],\; \bs{x}\in\Omega,
\end{equation*}
where $h(\bs{x},\bs{w})$ denotes the outcome of the \textit{input} and \textit{hidden} layers, being $\bs{w}$ the vector (matrix) of weight parameters, and $\sigma$ is the \textit{sigmoid} activation function defined as $\sigma(h)=(1+e^{-h})^{-1}$.

During the training procedure, the weights of the network are adjusted via \textit{backpropagation} in such a way that a certain objective function is minimized, i.e. we consider the following minimization problem
\begin{equation}\label{eq:obj_fun}
    \min_{\bs{w}}\ell(\hat{y}(\bs{x},\bs{w}),y)+\lambda R(\bs{w}),\; \bs{x}\in\Omega,
\end{equation}
where $\ell$ is a loss function and $R(\bs{w})$ is a possible regularization term that is controlled by a parameter $\lambda\in\mathbb{R}_{>0}$. The outcomes of this regression problem can be clustered by means of some thresholding process in order to construct a classifier whose predictive effectiveness is assessed by means of specific skill scores. We now introduce a one-parameter family of CMs from which we will derive a set of probabilistic skill scores. This approach will allow the introduction of a corresponding set of probabilistic loss functions and, accordingly, an {\em{a priori}} optimization of the scores. 

\subsection{Confusion matrices with probabilistic thresholds}
Let us consider a batch of predictions-labels $\mathcal{S}=\mathcal{S}_n(\bs{w})=\{(\hat{y}(\bs{x}_i,\bs{w}),y_i)\}_{i=1,\dots,n}$, where $y_i\in\{0,1\}$ is the true label associated to the element $\bs{x}_i\in\mathcal{X}$ and $\hat{y}(\bs{x}_i,\bs{w})\in[0,1]$ is the prediction for $\bs{x}_i$ given by the model. Furthermore, let $n^+$ and $n^-$ be the number of elements in $\mathcal{S}_n(\bs{w})$ whose true class $y_i$, $i=1,\dots,n$, is the \textit{positive} ($\{y=1\}$) or the \textit{negative} ($\{y=0\}$), respectively. Therefore, $n^++n^-=n$.

Let $\tau\in\R,\;\tau\in(0,1),$ and let
\begin{equation*}
    \mathbbm{1}_{x}(\tau)=\mathbbm{1}_{\{x>\tau\}}=\begin{dcases} 0 & \textrm{if } x\le \tau,\\
        1 & \textrm{if }x>\tau,\end{dcases}\quad x\in(0,1),
\end{equation*}
be the indicator function. The classical confusion matrix is defined as
\begin{equation*}
\mathrm{CM}(\tau)=\mathrm{CM}(\tau,\mathcal{S}_n(\bs{w})) = 
\begin{pmatrix}
\mathrm{TN}(\tau,\mathcal{S}_n(\bs{w})) & \mathrm{FP}(\tau,\mathcal{S}_n(\bs{w})) \\
\mathrm{FN}(\tau,\mathcal{S}_n(\bs{w})) & \mathrm{TP}(\tau,\mathcal{S}_n(\bs{w}))
\end{pmatrix},
\end{equation*}
where the elements are
\begin{equation}\label{eq:mat_defi}
    \begin{split}
         & \mathrm{TN}(\tau,\mathcal{S}_n(\bs{w})) = \sum_{i=1}^n{(1-y_i)\mathbbm{1}_{\{\hat{y}(\bs{x}_i,\bs{w})<\tau\}}},\;\mathrm{TP}(\tau,\mathcal{S}_n(\bs{w})) = \sum_{i=1}^n{y_i\mathbbm{1}_{\{\hat{y}(\bs{x}_i,\bs{w})>\tau\}}},\vspace{5pt}\\
         & \mathrm{FP}(\tau,\mathcal{S}_n(\bs{w})) = \sum_{i=1}^n{(1-y_i)\mathbbm{1}_{\{\hat{y}(\bs{x}_i,\bs{w})>\tau\}}},\;\mathrm{FN}(\tau,\mathcal{S}_n(\bs{w})) = \sum_{i=1}^n{y_i\mathbbm{1}_{\{\hat{y}(\bs{x}_i,\bs{w})<\tau\}}}.
        \end{split}
\end{equation}
In what follows, we let $\tau$ be a continuous random variable whose probability density function (pdf) $f$ is supported in $[a,b]\subseteq[0,1]$, $a,b\in\R$. Furthermore, we denote as $F$ the cumulative density function (cdf)
\begin{equation*}
    F(x)=\int_{a}^{x}{f(\xi)\mathrm{d}\xi},\quad x\le b.
\end{equation*}
We recall that
\begin{equation*}
    \mathbb{E}_{\tau}[\mathbbm{1}_{x}(\tau)]=\int_{a}^{b}{\mathbbm{1}_{\{x>\xi\}}f(\xi)\mathrm{d}\xi}=\int_{a}^{x}{f(\xi)\mathrm{d}\xi}=F(x).
\end{equation*}
This leads to the following definition.
\begin{definition}
Let $\mathcal{S}_n(\bs{w})$ be a batch of predictions-labels and let $\mathrm{CM}(\tau,\mathcal{S}_n(\bs{w}))$ be a confusion matrix as defined in \eqref{eq:mat_defi}. Moreover, let $\tau$ be a continuous random variable on $(0,1)$ with cdf $F$. We define the expected confusion matrix as
\begin{equation*}
\overline{\mathrm{CM}}_{F}=\overline{\mathrm{CM}}_F(\mathcal{S}_n(\bs{w}))=
\begin{pmatrix}
\overline{\mathrm{TN}}_F(\mathcal{S}_n(\bs{w})) & \overline{\mathrm{FP}}_F(\mathcal{S}_n(\bs{w})) \\
\overline{\mathrm{FN}}_F(\mathcal{S}_n(\bs{w})) & \overline{\mathrm{TP}}_F(\mathcal{S}_n(\bs{w}))
\end{pmatrix},
\end{equation*}
where
\begin{equation}\label{eq:fcm_entries}
    \begin{split}
         & \overline{\mathrm{TN}}_F(\mathcal{S}_n(\bs{w})) = \sum_{i=1}^n{(1-y_i)(1-F(\hat{y}(\bs{x}_i,\bs{w})))},\; \overline{\mathrm{TP}}_F(\mathcal{S}_n(\bs{w})) = \sum_{i=1}^n{y_iF(\hat{y}(\bs{x}_i,\bs{w}))},\vspace{5pt}\\
         & \overline{\mathrm{FP}}_F(\mathcal{S}_n(\bs{w})) = \sum_{i=1}^n{(1-y_i)F(\hat{y}(\bs{x}_i,\bs{w}))},\vspace{5pt}\; \overline{\mathrm{FN}}_F(\mathcal{S}_n(\bs{w})) = \sum_{i=1}^n{y_i(1-F(\hat{y}(\bs{x}_i,\bs{w})))}.
        \end{split}
\end{equation}
\end{definition}
\begin{remark}
We observe that $\overline{\mathrm{CM}}_F$ is indeed the expected value of the matrix $\mathrm{CM}(\tau)$ with respect to $\tau$, meaning that
\begin{equation}
    \mathbb{E}_{\tau}[\mathrm{CM}(\tau)]\coloneqq \begin{pmatrix}
\mathbb{E}_{\tau}[\mathrm{TN}(\tau,\mathcal{S}_n(\bs{w}))] & \mathbb{E}_{\tau}[\mathrm{FP}(\tau,\mathcal{S}_n(\bs{w}))] \\
\mathbb{E}_{\tau}[\mathrm{FN}(\tau,\mathcal{S}_n(\bs{w}))] & \mathbb{E}_{\tau}[\mathrm{TP}(\tau,\mathcal{S}_n(\bs{w}))]
\end{pmatrix}=\overline{\mathrm{CM}}_F.
\end{equation}
Moreover, the sum by rows is preserved, that is for any $\tau\in(0,1)$
\begin{equation*}
    \big\lVert \overline{\mathrm{CM}}_F(\mathcal{S}_n(\bs{w})) \big\lVert_{\infty}=\big\lVert \mathrm{CM}(\tau,\mathcal{S}_n(\bs{w})) \big\lVert_{\infty}=\max\{n^-,n^+\}.
\end{equation*}
Hence, all considered matrices are contained in the set
\begin{equation}\label{eq:def_c}
    \mathcal{C}(\mathcal{S}_n)\coloneqq \{\mathrm{A}\in \mathcal{M}^+_{2}\:|\: \mathrm{A}_{11}+\mathrm{A}_{12} = n^-,\;\mathrm{A}_{21}+\mathrm{A}_{22} = n^+\},
\end{equation}
where $\mathrm{A}=(\mathrm{A}_{ij})_{i,j=1,2}$ and $\mathcal{M}^+_{2}$ is the set of $2\times 2$ positive-valued matrices.
\end{remark}
Therefore, we can prove the following results by considering well-known concentration inequalities:
\begin{proposition}\label{prp:expected_matrix}
Let $\mathcal{S}_n\coloneqq\mathcal{S}_n(\bs{w})$ be a batch of predictions-labels, $n\in\mathbb{N}$, and let $\tau$ be a continuous random variable on $(0,1)$ with cdf $F$. Then,
letting $\mathrm{N}\in\{\mathrm{TN},\mathrm{FP}\}$, $\mathrm{P}\in\{\mathrm{TP},\mathrm{FN}\}$ and $\varepsilon>0$, we have
\begin{equation}\label{eq:thx_mc}
    \begin{split}
        & \mathbb{P}(|\mathrm{N}(\tau,\mathcal{S}_n)-\overline{\mathrm{N}}_F(\mathcal{S}_n)|\ge\varepsilon)\le 2\: \mathrm{exp}\bigg(-\frac{2\varepsilon^2}{(n^-)^2}\bigg),\\
        & \mathbb{P}(|\mathrm{P}(\tau,\mathcal{S}_n)-\overline{\mathrm{P}}_F(\mathcal{S}_n)|\ge\varepsilon)\le 2\: \mathrm{exp}\bigg(-\frac{2\varepsilon^2}{(n^+)^2}\bigg).  
    \end{split}
\end{equation}
Moreover, denoting as $\mathrm{tr}(\mathrm{CM})=\mathrm{TN}+\mathrm{TP}$ the trace of the confusion matrix, we get
\begin{equation}\label{eq:thx_ho}
    \mathbb{P}(|\mathrm{tr}(\mathrm{CM}(\tau,\mathcal{S}_n)) - \mathrm{tr}(\overline{\mathrm{CM}}_F(\mathcal{S}_n))|\ge\varepsilon)\le 2\: \mathrm{exp}\bigg(-\frac{2\varepsilon^2}{(n^-)^2+(n^+)^2}\bigg).
\end{equation}
\end{proposition}
\begin{proof}
In order to prove \eqref{eq:thx_mc} it is sufficient to consider the McDiarmid's inequality (see e.g. \cite[Theorem 4.5, p.88]{ShaweTaylor04}) and to observe that
\begin{equation*}
    \sup_{\tau_1,\tau_2\in(0,1)}{|\mathrm{N}(\tau_1,\mathcal{S}_n)-\mathrm{N}(\tau_2,\mathcal{S}_n)|}=n^-,\; \sup_{\tau_1,\tau_2\in(0,1)}{|\mathrm{P}(\tau_1,\mathcal{S}_n)-\mathrm{P}(\tau_2,\mathcal{S}_n)|}=n^+.
\end{equation*}
Then, we can apply the Hoeffding's inequality \cite{Hoeffding63} to $\mathrm{TN}(\tau,\mathcal{S}_n)+\mathrm{TP}(\tau,\mathcal{S}_n)$,  thus obtaining \eqref{eq:thx_ho}.
\end{proof}

\subsection{Derived scores and loss functions}

A skill score is a function mapping the space of confusion matrices onto $\R$. In the classical setting, the skill score is denoted as $s:M_{2,2}(\N) \rightarrow \R$ and is constructed on $\mathrm{CM}(\tau,\mathcal{S}_n(\bs{w}))$ for a fixed value of $\tau \in (0,1)$. In the case of confusion matrices with probabilistic threshold, the score explicitly depends on the cumulative distribution function $F$ adopted and is therefore denoted as ${\overline{s}}_F:M_{2,2}(\R) \rightarrow \R$. In this latter case, we can introduce a class of Score-Oriented Loss (SOL) functions as follows.
\begin{definition}\label{def:scoloss}
Let $\mathcal{S}_n(\bs{w})$ be a batch of predictions-labels and
let $s(\tau,\mathcal{S}_n(\bs{w}))=s(\tau)$ be a score calculated upon $\mathrm{CM}(\tau,\mathcal{S}_n(\bs{w}))$, $\tau\in(0,1)$. Indicating with ${\overline{s}}_F(\mathcal{S}_n(\bs{w}))={\overline{s}}_F$ the score $s$ calculated upon $\overline{\mathrm{CM}}_F(\mathcal{S}_n(\bs{w}))$, a Score-Oriented Loss (SOL) function related to $s$ is defined as
\begin{equation*}
    \ell_{\overline{s}}(F,\mathcal{S}_n(\bs{w})))\coloneqq - {\overline{s}}_F(\mathcal{S}_n(\bs{w}))).
\end{equation*}
\end{definition}
As an example, taking the \textit{accuracy} score
\begin{equation*}
    \textrm{acc}=\frac{\textrm{TP}+\textrm{TN}}{\textrm{TP}+\textrm{TN}+\textrm{FP}+\textrm{FN}},
\end{equation*}
we get
\begin{equation*}
\begin{split}
    \overline{\textrm{acc}}_F(\mathcal{S}_n(\bs{w}))) & \coloneqq \frac{\sum_{i=1}^n{y_iF(\hat{y}(\bs{x}_i,\bs{w}))+(1-y_i)(1-F(\hat{y}(\bs{x}_i,\bs{w})))}}{n}
\end{split}
\end{equation*}
and then the loss function $\ell_{\overline{\textrm{acc}}}(F,\mathcal{S}_n(\bs{w}))\coloneqq - \overline{\textrm{acc}}_F(\mathcal{S}_n(\bs{w})))$.

\begin{remark}
By setting $0/0\coloneqq 0$ and by virtue of the fundamental theorem of calculus, we observe that $\ell_{\overline{s}}$ is derivable with respect to the weights vector $\bs{w}$, which is a fundamental property for its usage as loss function.
\end{remark}

We now prove the following

\begin{theorem}\label{thm:linea}
Let $\mathcal{S}_n\coloneqq\mathcal{S}_n(\bs{w})$ be a batch of predictions-labels, $n\in\mathbb{N}$, and let $\tau$ be a continuous random variable on $(0,1)$ with cdf $F$. Then, if $s$ is linear with respect to the entries of the confusion matrix, we have
\begin{equation}\label{eq:score_expected}
    \mathbb{E}_{\tau}[s(\tau,\mathcal{S}_n)]=\overline{s}_F(\mathcal{S}_n).
\end{equation}
Moreover, letting $\mathrm{supp}(s(\tau,\mathcal{S}_n))=[a,b]$, $a,b\in\mathbb{R}$, $a<b$, and $\varepsilon>0$, we obtain
\begin{equation}\label{eq:score_proba}
    \mathbb{P}(|s(\tau,\mathcal{S}_n)-\overline{s}_F(\mathcal{S}_n)|\ge\varepsilon)\le 2\: \mathrm{exp}\bigg(-\frac{2\varepsilon^2}{(b-a)^2}\bigg).
\end{equation}
\end{theorem}
\begin{proof}
Due to the linearity of the expected value, the results in Proposition \ref{prp:expected_matrix} imply \eqref{eq:score_expected}. As a consequence, the McDiarmid's inequality can be applied in order to obtain \eqref{eq:score_proba}.
\end{proof}
\begin{remark}\label{rem:rems}
Suppose that the score is not linear with respect to the entries of the vectorized confusion matrix $\mathrm{vec}(\mathrm{CM}(\tau))=(\mathrm{TN}(\tau),\mathrm{FP}(\tau),\mathrm{FN}(\tau),\mathrm{TP}(\tau))$. By looking at the scores $s$ and $\overline{s}_F$ as functions of $\mathrm{vec}(\mathrm{CM}(\tau))$, i.e. $s(\tau)=s(\mathrm{vec}(\mathrm{CM}(\tau)))$ and $\overline{s}_F=s(\mathrm{vec}(\overline{\mathrm{CM}}_F))$, we have the Taylor approximation (see e.g. \cite{Benaroya05})
\begin{equation*}
    s(\mathrm{vec}(\mathrm{CM}(\tau)))\approx s(\mathrm{vec}(\overline{\mathrm{CM}}_F))+\sum_{|\bs{\alpha}|\ge 1}{\frac{D^{\bs{\alpha}}s(\mathrm{vec}(\overline{\mathrm{CM}}_F))}{\bs{\alpha}!}(\mathrm{vec}(\mathrm{CM}(\tau)-\overline{\mathrm{CM}}_F))^{\bs{\alpha}}},
\end{equation*}
which implies
\begin{equation*}
    \mathbb{E}_{\tau}[s(\tau)]\approx \overline{s}_F+\sum_{|\bs{\alpha}|\ge 1}{\frac{D^{\bs{\alpha}}\overline{s}_F}{\bs{\alpha}!}\mathbb{E}_{\tau}[(\mathrm{vec}(\mathrm{CM}(\tau)-\overline{\mathrm{CM}}_F))^{\bs{\alpha}}]},
\end{equation*}
where $\bs{\alpha}\in\mathbb{N}^4$ denotes the classical multi-index notation.
\end{remark}

Recalling \eqref{eq:obj_fun}, let $T(\bs{w})\coloneqq \ell(\mathcal{S}(\bs{w}))+\lambda R(\bs{w})$ be the target function to minimize during the training of the network, where $\ell$ is a generic loss function. When performing \textit{a posteriori} maximization of a score $s$, letting $\bs{w}^{\star}=\argmin_{\bs{w}}T(\bs{w})$, we compute
\begin{equation*}
    \max_{\tau\in(0,1)}s(\tau,\mathcal{S}(\bs{w}^{\star})).
\end{equation*}
In our setting with SOL functions, in view of Definition \ref{def:scoloss} and Remark \ref{rem:rems}, the training minimization problem results in
\begin{equation*}
    \min_{\bs{w}}\ell_{\overline{s}}(F,\mathcal{S}(\bs{w})))+\lambda R(\bs{w})\approx \max_{\bs{w}} \mathbb{E}_{\tau}[s(\tau,\mathcal{S}(\bs{w}))]-\lambda R(\bs{w}),
\end{equation*}
where the equality is achieved under the assumptions of Theorem \ref{thm:linea}. Therefore, the maximization of the expected value of the score is included in the training process. Moreover, we observe that while a direct maximization of the score $s(\tau^{\star},\mathcal{S}(\bs{w}))$ for a fixed $\tau^{\star}\in(0,1)$ is not possible due to the lack of derivability, this procedure can be approximated by considering for $\tau$ a probability distribution highly concentrated around the mean value $\tau^{\star}$. In Section \ref{sec:results}, we review two different probability distributions that express different concentration properties.

The following result concerns the mean absolute deviation (mad) \cite{Geary35} of a Lipschitz continuous score $s$ from its expected value $\overline{s}_F$.
\begin{theorem}\label{thm:lippa}
Let $\mathcal{S}_n$ be a batch of predictions-labels, $n\in\mathbb{N}$, and let $\tau$ be a continuous random variable on $(0,1)$ with cdf $F$. Assume that $s$ is a Lipschitz continuous function with respect to the entries of  $\mathrm{CM}(\tau,\mathcal{S}_n)$ on the set $\mathcal{C}(\mathcal{S}_n)$. Then,
\begin{equation*}
    \mathbb{E}_{\tau}[|s(\tau,\mathcal{S}_n)-{\overline{s}}_F(\mathcal{S}_n)|]\le \frac{1}{2}\: K_{s}(J_F+\mathrm{mad}_{\tau}[\mathrm{TN}(\tau)]+\mathrm{mad}_{\tau}[\mathrm{TP}(\tau)]),
\end{equation*}
where
\begin{equation}\label{eq:lip_cons}
    K_s\coloneqq \sup_{\mathrm{A}\in\mathcal{C}(\mathcal{S}_n)}\lVert \nabla s(\mathrm{A}_{11},\mathrm{A}_{12},\mathrm{A}_{21},\mathrm{A}_{22}) \lVert_{\infty},
\end{equation}
with $\mathcal{C}(\mathcal{S}_n)$ as defined in \eqref{eq:def_c}, 
\begin{equation}\label{eq:weird_cons}
    J_F\coloneqq \mathbb{E}_{\tau}[\:|\:|\mathrm{TN}(\tau)-\overline{\mathrm{TN}}_F|-|\mathrm{TP}(\tau)-\overline{\mathrm{TP}}_F|\:|\:],
\end{equation}
and $mad$ indicates the mean absolute deviation.
\end{theorem}
\begin{proof}
As in Remark \ref{rem:rems}, we consider $s$ as a function of the vectorized confusion matrix $\mathrm{vec}(\mathrm{CM}(\tau))=(\mathrm{TN}(\tau),\mathrm{FP}(\tau),\mathrm{FN}(\tau),\mathrm{TP}(\tau))$, and thus
\begin{equation*}
{\overline{s}}_F=s(\overline{\mathrm{TN}}_F,\overline{\mathrm{FP}}_F,\overline{\mathrm{FN}}_F,\overline{\mathrm{TP}}_F).
\end{equation*}
The Lipschitz continuity of the score $s$ yields to
\begin{equation}\label{eq:lippi}
\begin{split}
     &|s(\mathrm{TN}(\tau),\mathrm{FP}(\tau),\mathrm{FN}(\tau),\mathrm{TP}(\tau))-s(\overline{\mathrm{TN}}_F,\overline{\mathrm{FP}}_F,\overline{\mathrm{FN}}_F,\overline{\mathrm{TP}}_F)|\le\\ 
     & K_{s}\lVert (\mathrm{TN}(\tau),\mathrm{FP}(\tau),\mathrm{FN}(\tau),\mathrm{TP}(\tau))-(\overline{\mathrm{TN}}_F,\overline{\mathrm{FP}}_F,\overline{\mathrm{FN}}_F,\overline{\mathrm{TP}}_F)\lVert_{\infty},
\end{split}
\end{equation}
where $K_{s}$ is the Lipschitz constant defined in \eqref{eq:lip_cons}. Then, since
\begin{equation*}
\mathrm{CM}(\tau,\mathcal{S}_n),\overline{\mathrm{CM}}_F(\mathcal{S}_n)\in\mathcal{C}(\mathcal{S}_n),
\end{equation*}
we have
\begin{equation*}
\begin{split}
    & \lVert (\mathrm{TN}(\tau),\mathrm{FP}(\tau),\mathrm{FN}(\tau),\mathrm{TP}(\tau))-(\overline{\mathrm{TN}}_F,\overline{\mathrm{FP}}_F,\overline{\mathrm{FN}}_F,\overline{\mathrm{TP}}_F)\lVert_{\infty}=\\
    & \max\{|\mathrm{TN}(\tau)-\overline{\mathrm{TN}}_F|,|\mathrm{TP}(\tau)-\overline{\mathrm{TP}}_F|\}=\\
    & \frac{|\mathrm{TN}(\tau)-\overline{\mathrm{TN}}_F|+|\mathrm{TP}(\tau)-\overline{\mathrm{TP}}_F|+|\:|\mathrm{TN}(\tau)-\overline{\mathrm{TN}}_F|-|\mathrm{TP}(\tau)-\overline{\mathrm{TP}}_F|\:|}{2}. 
    \end{split}
\end{equation*}
We proceed by taking the expected value of both sides of \eqref{eq:lippi}. Then, in order to conclude the proof, it is sufficient to observe that
\begin{equation*}
    \mathbb{E}_{\tau}[|\mathrm{TN}(\tau)-\overline{\mathrm{TN}}_F|]=\mathrm{mad}_{\tau}[\mathrm{TN}(\tau)],\quad     \mathbb{E}_{\tau}[|\mathrm{TP}(\tau)-\overline{\mathrm{TP}}_F|]=\mathrm{mad}_{\tau}[\mathrm{TP}(\tau)],
\end{equation*}
and to define $J_F$ as in \eqref{eq:weird_cons}.
\end{proof}
\begin{remark}
Concerning the bound presented in Theorem \ref{thm:lippa}, we observe that the Lipschitz constant $K_s$ is decreasing as $n$ gets larger. Indeed, the larger the number of samples in the batch, the smaller the variation of the score under a little alteration of the entries of the confusion matrix.
For example, we have
\begin{equation*}
    \nabla \mathrm{acc}(\mathrm{TN},\mathrm{FP},\mathrm{FN},\mathrm{TP})=\frac{1}{n^2}(n-\mathrm{TN}-\mathrm{TP}, n-\mathrm{TN}-\mathrm{TP},-\mathrm{TN}-\mathrm{TP},-\mathrm{TN}-\mathrm{TP})
\end{equation*}
which yields to $K_{\mathrm{acc}}=1/n$.
\end{remark}


\section{Applications to experimental forecasting problems}\label{sec:results}

The implementation of a machine learning approach based on SOL functions requires first the choice of the probability distribution function associated to the threshold $\tau$. In this paper we will utilize two distributions supported in $[0,1]$. 

\begin{figure}[htbp]
    \centering
    \begin{subfigure}[t]{.4\linewidth}
    \centering
   \includegraphics[width=\linewidth]{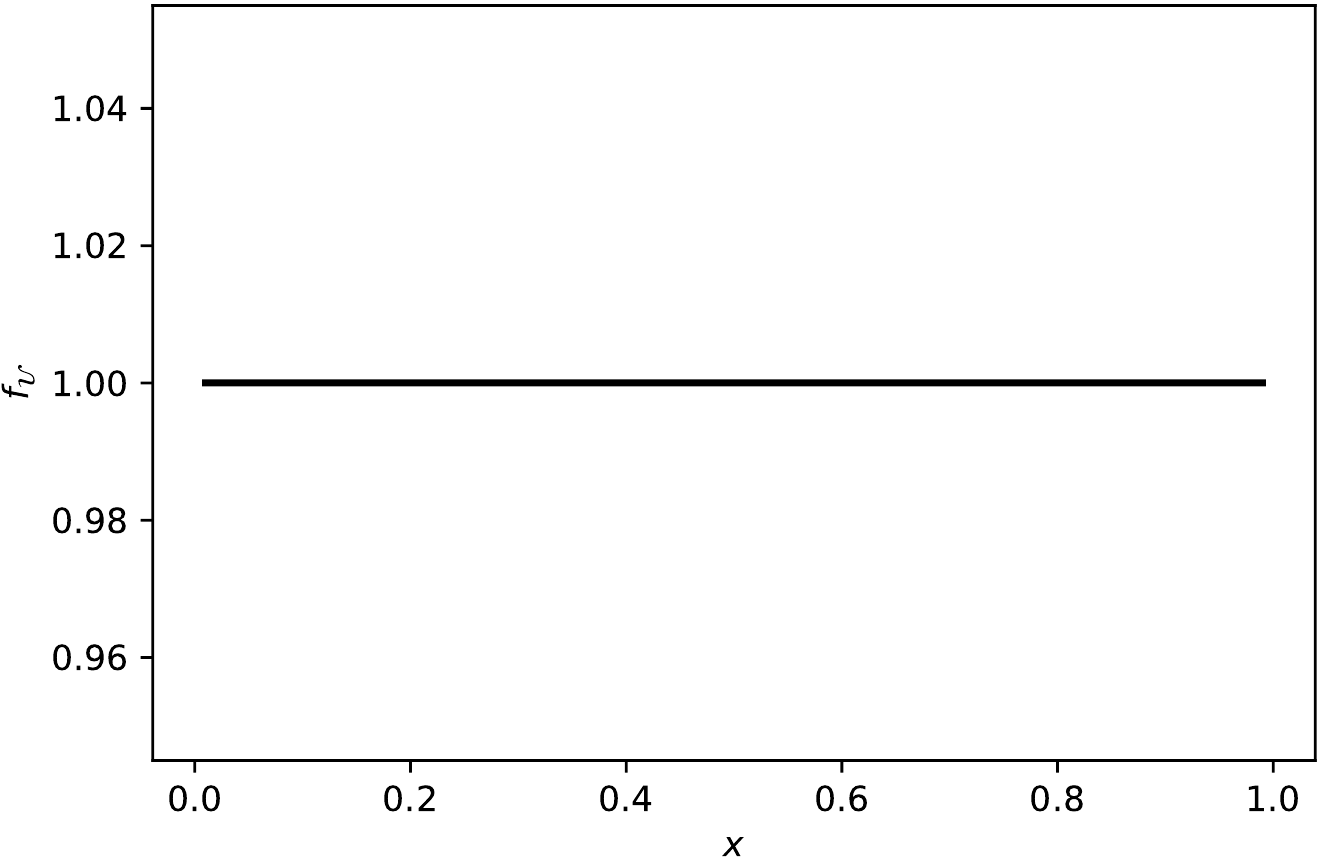}
    \caption{The function $f_{\mathcal{U}}$.}
    \end{subfigure}
    \begin{subfigure}[t]{.4\linewidth}
    \centering
   \includegraphics[width=\linewidth]{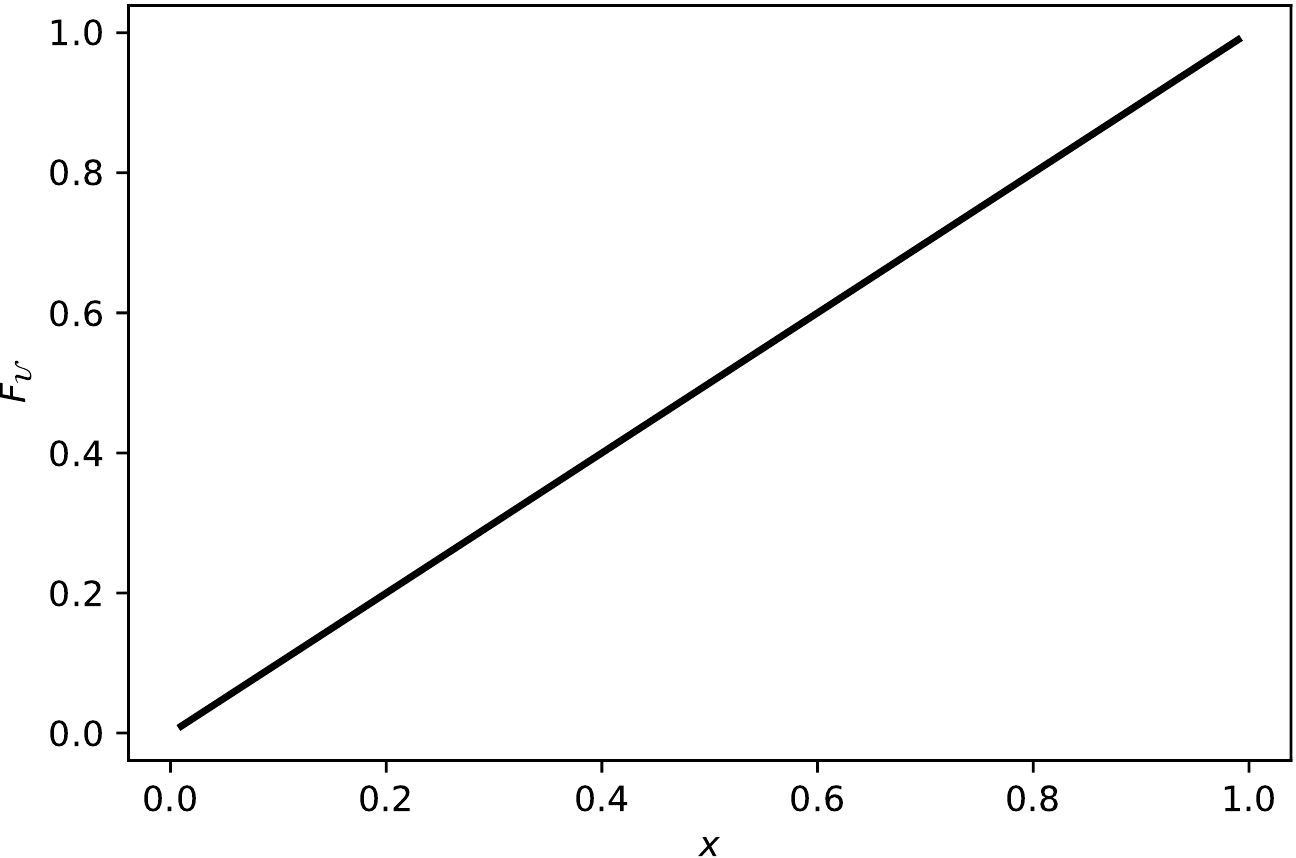} 
    \caption{The function $F_{\mathcal{U}}$.}
    \end{subfigure}
    \begin{subfigure}[t]{.4\linewidth}
    \centering
   \includegraphics[width=\linewidth]{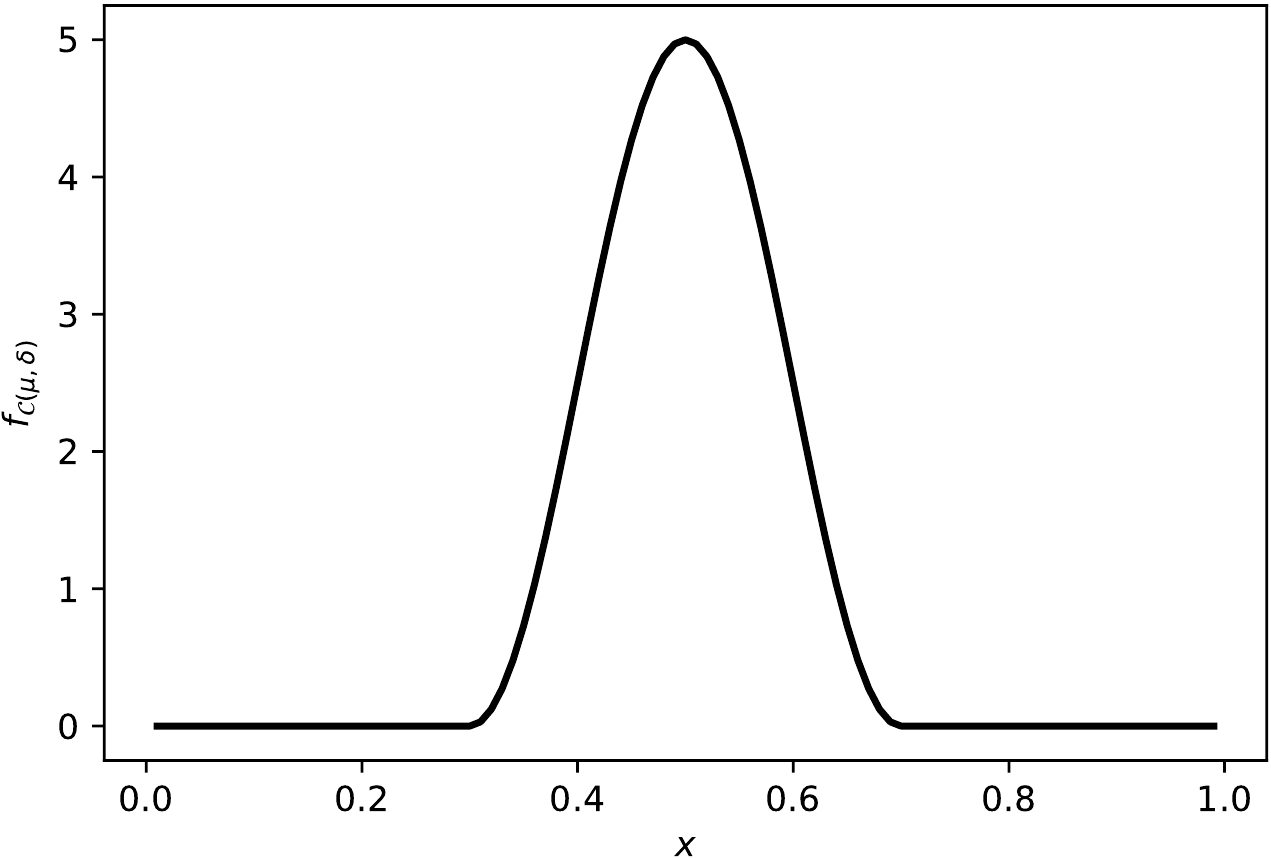} 
    \caption{The function $f_{\mathcal{C}(\mu,\delta)}$.}
    \end{subfigure}
    \begin{subfigure}[t]{.4\linewidth}
    \centering
   \includegraphics[width=\linewidth]{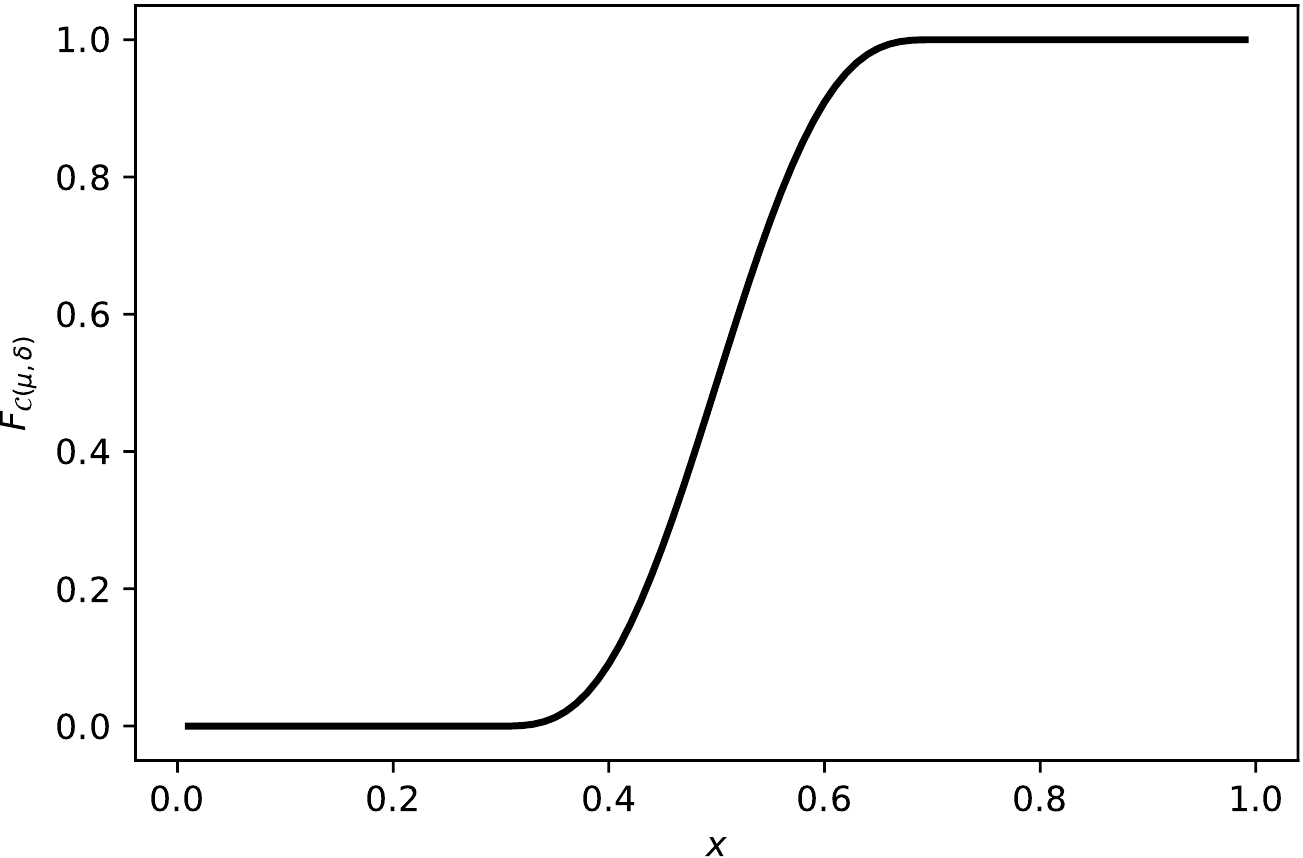} 
    \caption{The function $F_{\mathcal{C}(\mu,\delta)}$.}
    \end{subfigure}
    \caption{Top row: The functions $f_{\mathcal{U}}(x)$ (left panel) and $F_{\mathcal{U}}(x)$ (right panel), $x\in(0,1)$ for the uniform distribution. Bottom row: the functions $f_{\mathcal{C}(\mu,\delta)}(x)$ (left panel) and $F_{\mathcal{C}(\mu,\delta)}(x)$ (right panel), $x\in(0,1)$, for $\mu=0.5$ and $\delta=0.2$, for the raised cosine distribution.}
   \label{fig:distributions}
\end{figure}

Specifically, if $\tau\sim\mathcal{U}(0,1)$, i.e. it is a random variable uniformly distributed on $(0,1)$, its pdf and cdf are $f_{\mathcal{U}}(x)= 1$ and $F_{\mathcal{U}}(x)=x$ respectively, $x\in(0,1)$ (see Figure \ref{fig:distributions}, top row). This well-known distribution is characterized by a high variance and does not favor any particular value of the threshold. Therefore, with respect to a classical confusion matrix \eqref{eq:mat_defi}, in $\overline{\mathrm{CM}}_{F_{\mathcal{U}}}$ the indicator functions of the outputs are replaced by the outputs themselves. Although its probabilistic derivation from the uniform distribution is not trivial, from an empirical viewpoint the confusion matrix $\overline{\mathrm{CM}}_{F_{\mathcal{U}}}$ is probably the most straightforward way to extend a classical confusion matrix to having non-integer valued entries \cite{Lawson14}.

Let us now consider $\mu\in (0,1)$ and let $\delta\in\R,\;\delta>0$ be such that $[\mu-\delta,\mu+\delta]\subseteq[0,1]$. We take $\tau\sim\mathcal{C}(\mu,\delta)$, i.e. it is a random variable distributed according to the \textit{raised cosine} distribution, whose related pdf and cdf on (0,1) are
\begin{equation*}
\begin{split}
    & f_{\mathcal{C}(\mu,\delta)}(x)=\frac{1}{2\delta}\bigg(1+\cos\bigg(\frac{x-\mu}{\delta}\pi\bigg)\bigg)\mathbbm{1}_{\{\mu-\delta<x<\mu+\delta\}},\\
  & F_{\mathcal{C}(\mu,\delta)}(x)=\frac{1}{2}\bigg(1+\frac{x-\mu}{\delta}+ \frac{1}{\pi}\sin\bigg(\frac{x-\mu}{\delta}\pi\bigg)\bigg)\mathbbm{1}_{\{\mu-\delta<x<\mu+\delta\}}+\mathbbm{1}_{\{x\ge\mu+\delta\}}.
\end{split}
\end{equation*}
In this case $\mathbb{E}[\tau]=\mu$ and $\mathrm{Var}[\tau]=\delta^2(\pi^2-6)/(3\pi^2)$. We plotted $f_{\mathcal{C}(\mu,\delta)}$ and $F_{\mathcal{C}(\mu,\delta)}$ in Figure \ref{fig:distributions}, bottom row. Differently from the uniform case, the raised cosine distribution can potentially drive the model to be maximized for a restricted interval of threshold values around $\mu$, since we can achieve a high concentration by choosing a \textit{small} $\delta$.

We now test the performances of SOL functions relying on these two distribution functions, when applied to two actual binary classification problems.

A \textsc{Python} implementation of the SOLs is available for the scientific community at $\texttt{https://github.com/cesc14/SOL}\:.$

\subsection{Optimized \textit{a posteriori} thresholds and \textit{a priori} distributions}

We first considered the \textit{Adult} training dataset \cite{Kohavi96}, where the task is to predict whether a person can make more than 50,000 dollars per year based on census data. First, we performed the following pre-processing steps:
\begin{enumerate}
    \item
    We removed the samples containing missing values.
    \item
    We dropped the feature \textit{education}, which is highly correlated to the feature \textit{education-num}. Indeed, both features concern the education level of the people in the dataset.
    \item
    The possible values in the \textit{native-country} column, which reports the country of origin of the people in the data set, were restricted to $1$ in place of \textit{United-States} and $0$ for all other countries.
    \item
    All categorical features were \textit{one-hot} encoded. Finally, all features were standardized.
\end{enumerate}
The resulting elements were then described by $48$ numerical features. The data set consisted of 30162 samples, where 7508 people made more than 50,000 dollars per year, while 22654 did not. As for the network architecture, we used a fully-connected feed-forward NN with an input layer ($50$ ReLu neurons), two hidden layers ($20$ and $5$ ReLu neurons) and a sigmoid output unit. During the training phase, $1/3$ of training set was used as a validation set. We trained the network on $500$ epochs and we imposed an early stopping condition when no improvement in the validation loss was obtained after $30$ epochs. 

The following experiments were devoted to show how the choice of the distribution of the threshold $\tau$ affected the result of the \textit{a posteriori} score maximization performed by varying the threshold in $(0,1)$. In doing so, we considered the SOL related to the \textit{f$_1$-score} \cite{lipton2014optimal}, which is the harmonic mean of precision and recall, i.e.
\begin{equation*}
    \textrm{f$_1$-score}=\frac{2\textrm{TP}}{2\textrm{TP}+\textrm{FP}+\textrm{FN}}.
\end{equation*}
We observe that this score does not satisfy the linearity assumption in Theorem \ref{thm:linea} (see also Remark \ref{rem:rems}).

We tested the uniform distribution $\mathcal{U}$ as well as the raised cosine distribution $\mathcal{C}(\mu,\delta)$ with different values of the parameters $\mu,\delta$. More precisely, for each combination of $(\mu,\delta)\in \{0.3,0.5,0.7\}\times \{0.1,0.2,0.3\}$ 
we randomly selected $4/5$ elements in the data set for training the NN
and we performed the \textit{a posteriori} score maximization on the training set.
    We repeated this procedure $100$ times.

In Table \ref{tab:prima}, we present the obtained results. We report that in some cases the training optimization process was unsuccessful, that is the validation loss got stuck from the beginning of the training, and no actual learning took place. Therefore, we did not include such cases in the computation of the other entries of the table.

\begin{figure}[htbp]
  \centering
  $$
  \begin{array}{c}
    \begin{subfigure}[t]{.4\linewidth}
    \centering
    \includegraphics[width=\linewidth]{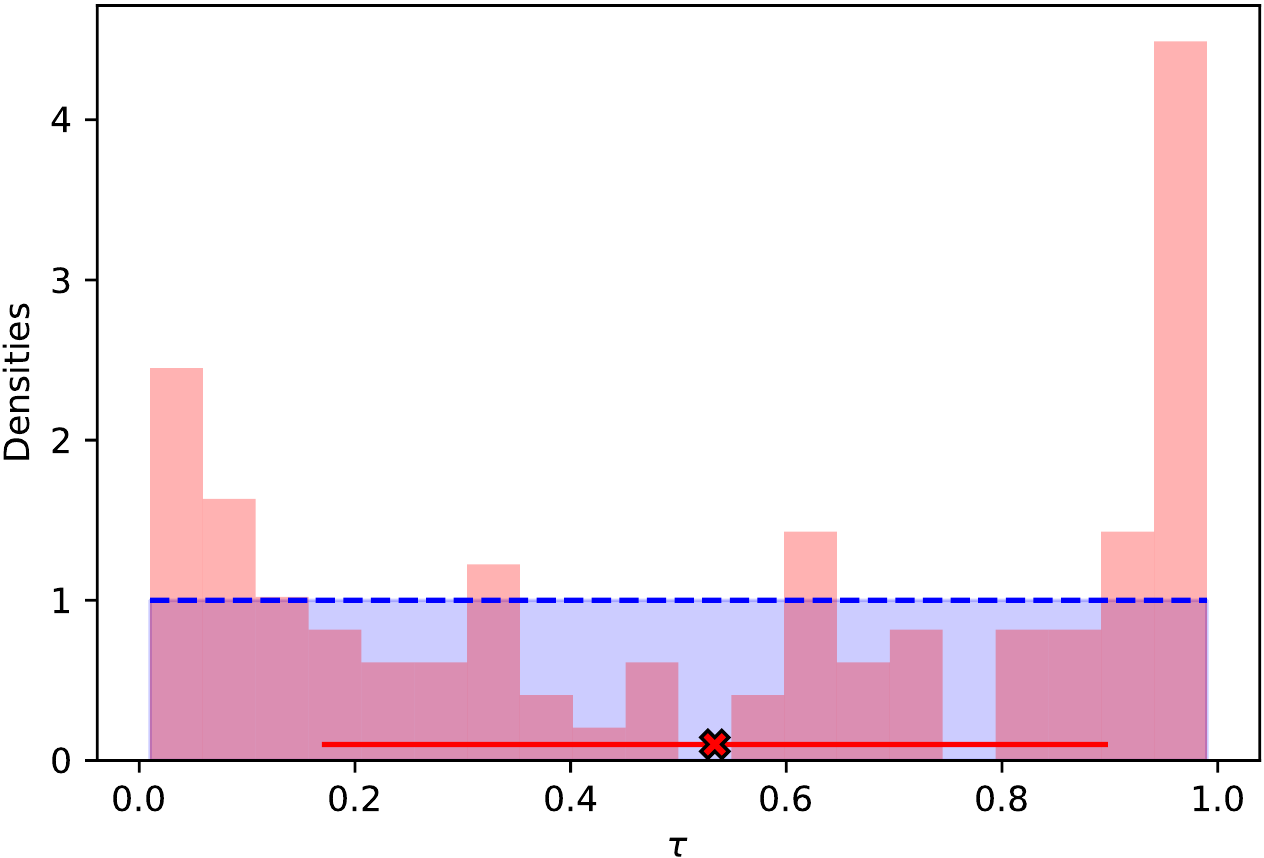}
    \caption{Uniform}
  \end{subfigure}\\
    \begin{subfigure}[t]{.4\linewidth}
    \centering
   \includegraphics[width=\linewidth]{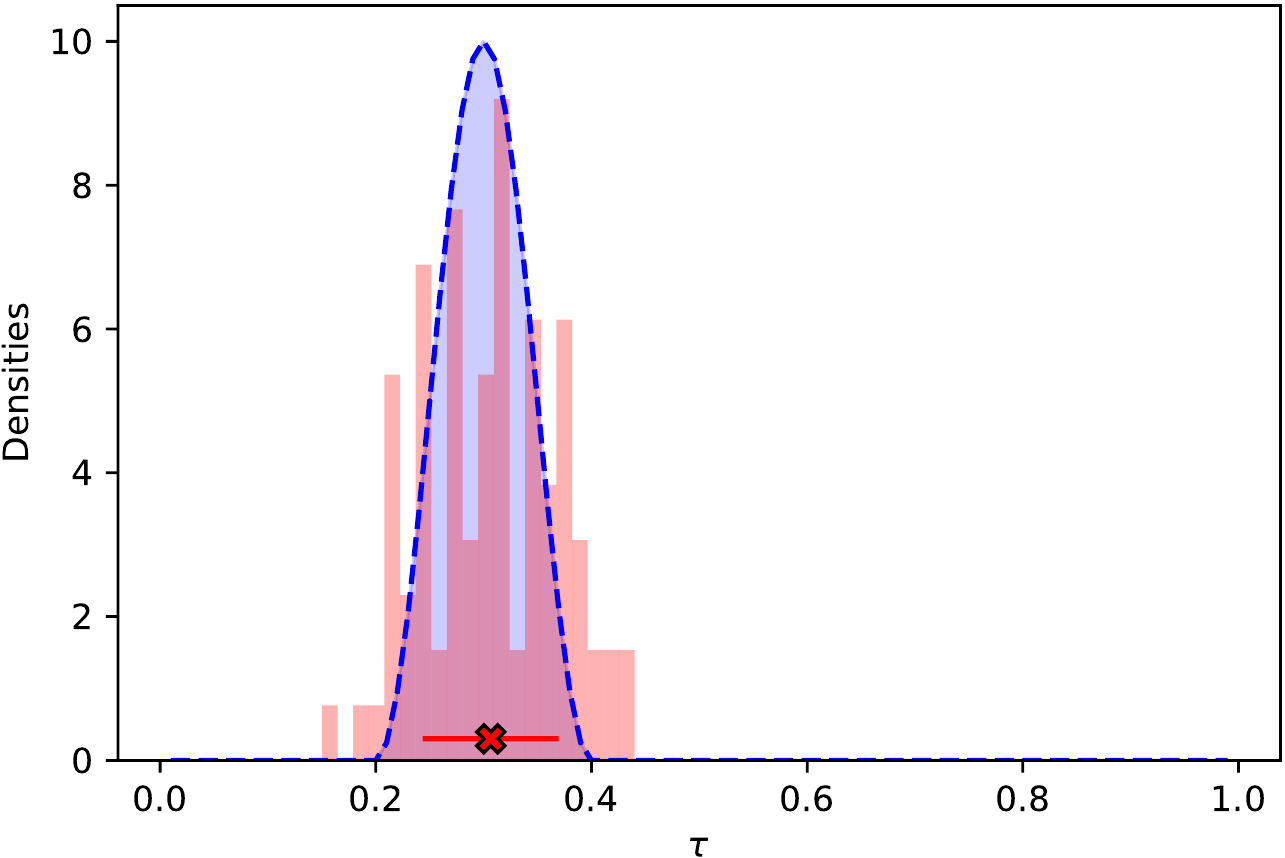} 
    \caption{Raised cosine: $\mu=0.3$, $\delta=0.1$.}
  \end{subfigure}
    \begin{subfigure}[t]{.4\linewidth}
    \centering
    \includegraphics[width=\linewidth]{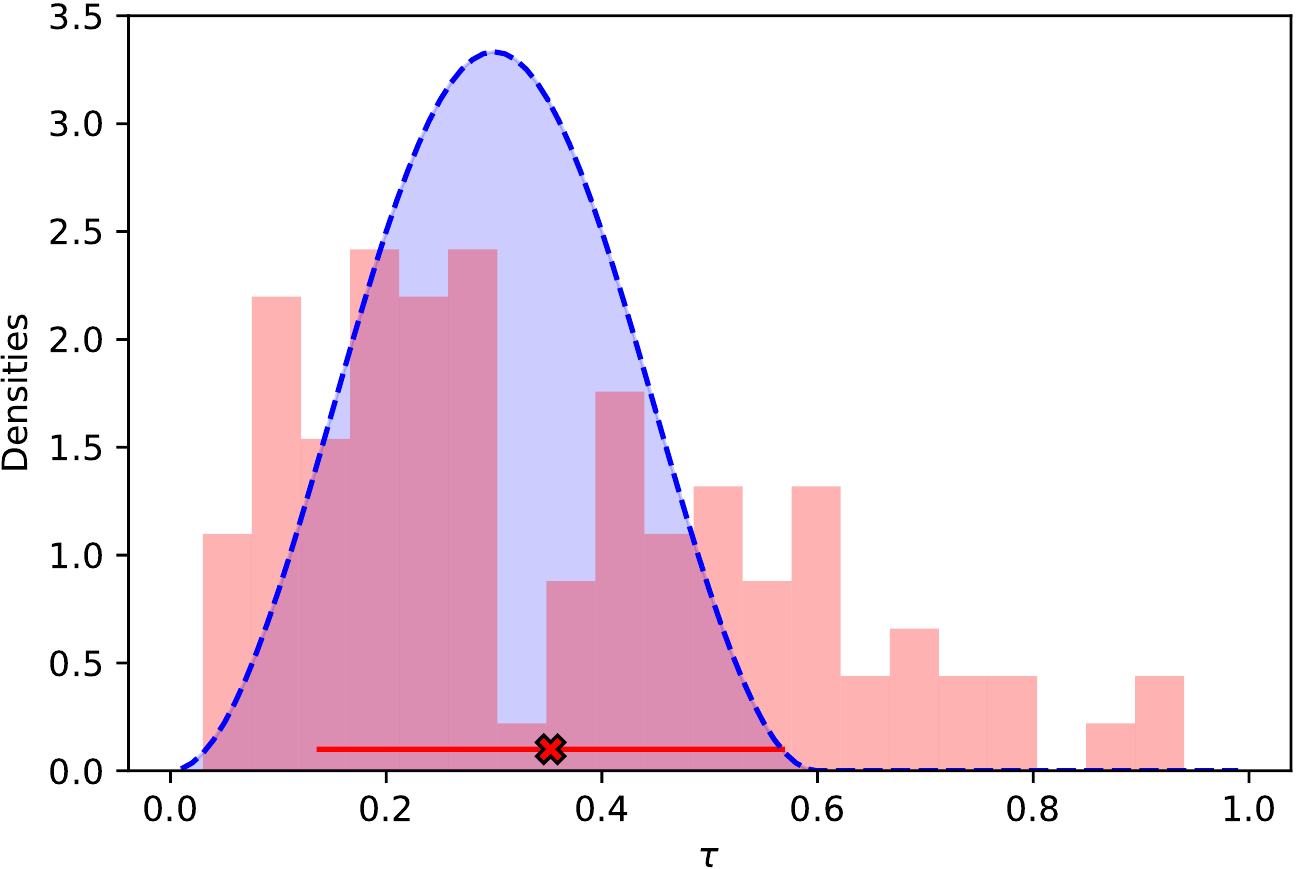}
    \caption{Raised cosine: $\mu=0.3$, $\delta=0.3$.}
  \end{subfigure}\\
    \begin{subfigure}[t]{.4\linewidth}
    \centering
   \includegraphics[width=\linewidth]{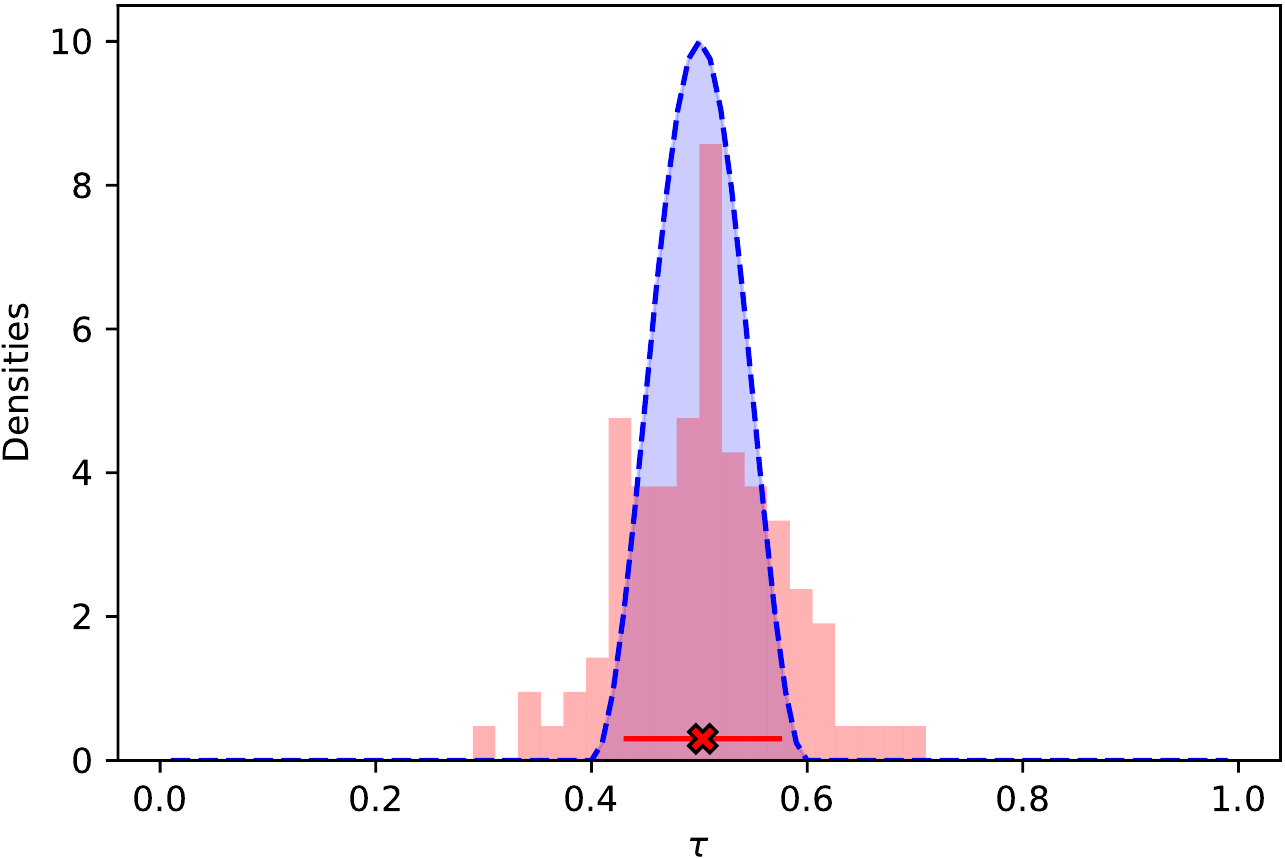} 
    \caption{Raised cosine: $\mu=0.5$, $\delta=0.1$.}
  \end{subfigure}
    \begin{subfigure}[t]{.4\linewidth}
    \centering
    \includegraphics[width=\linewidth]{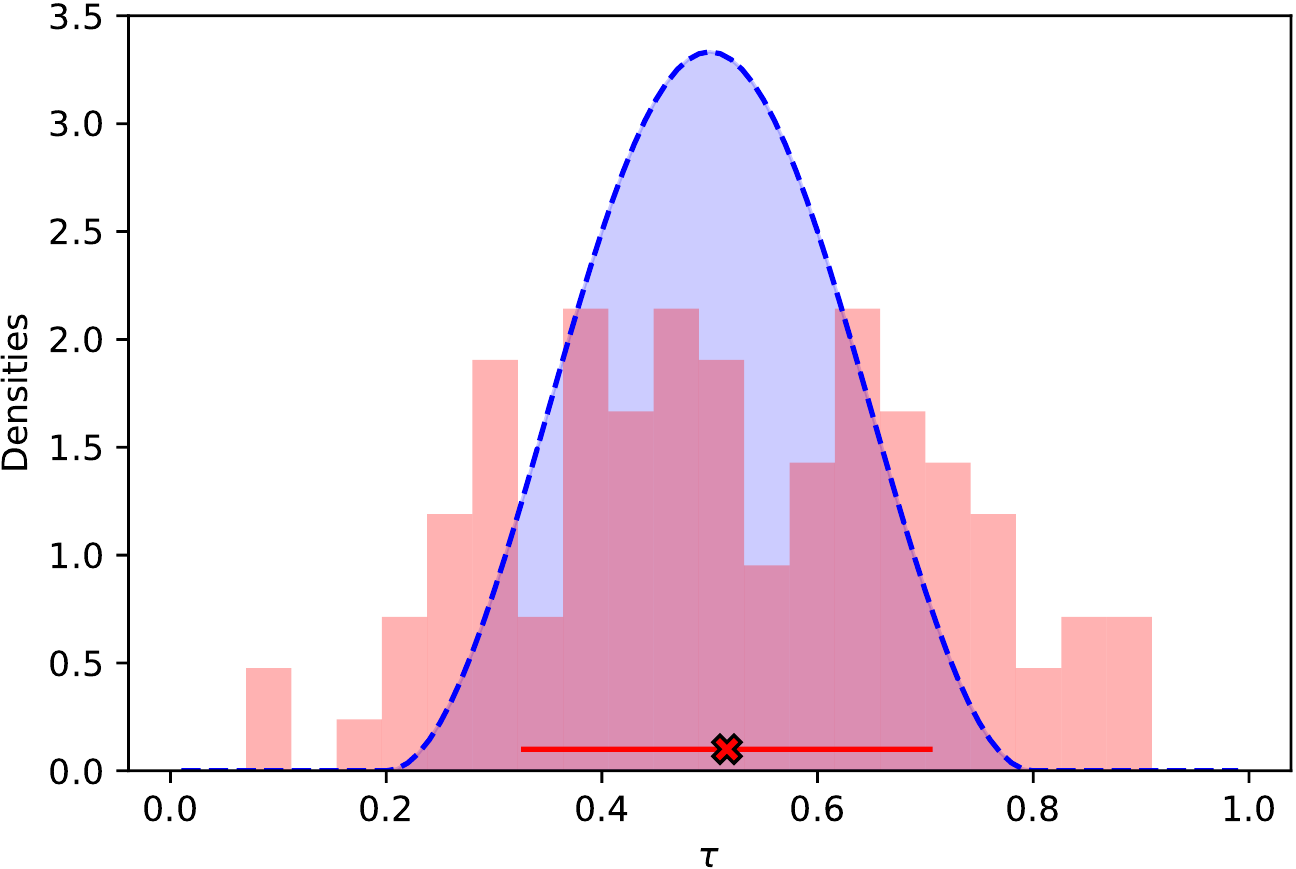}
    \caption{Raised cosine: $\mu=0.5$, $\delta=0.3$.}
  \end{subfigure}\\  
    \begin{subfigure}[t]{.4\linewidth}
    \centering
   \includegraphics[width=\linewidth]{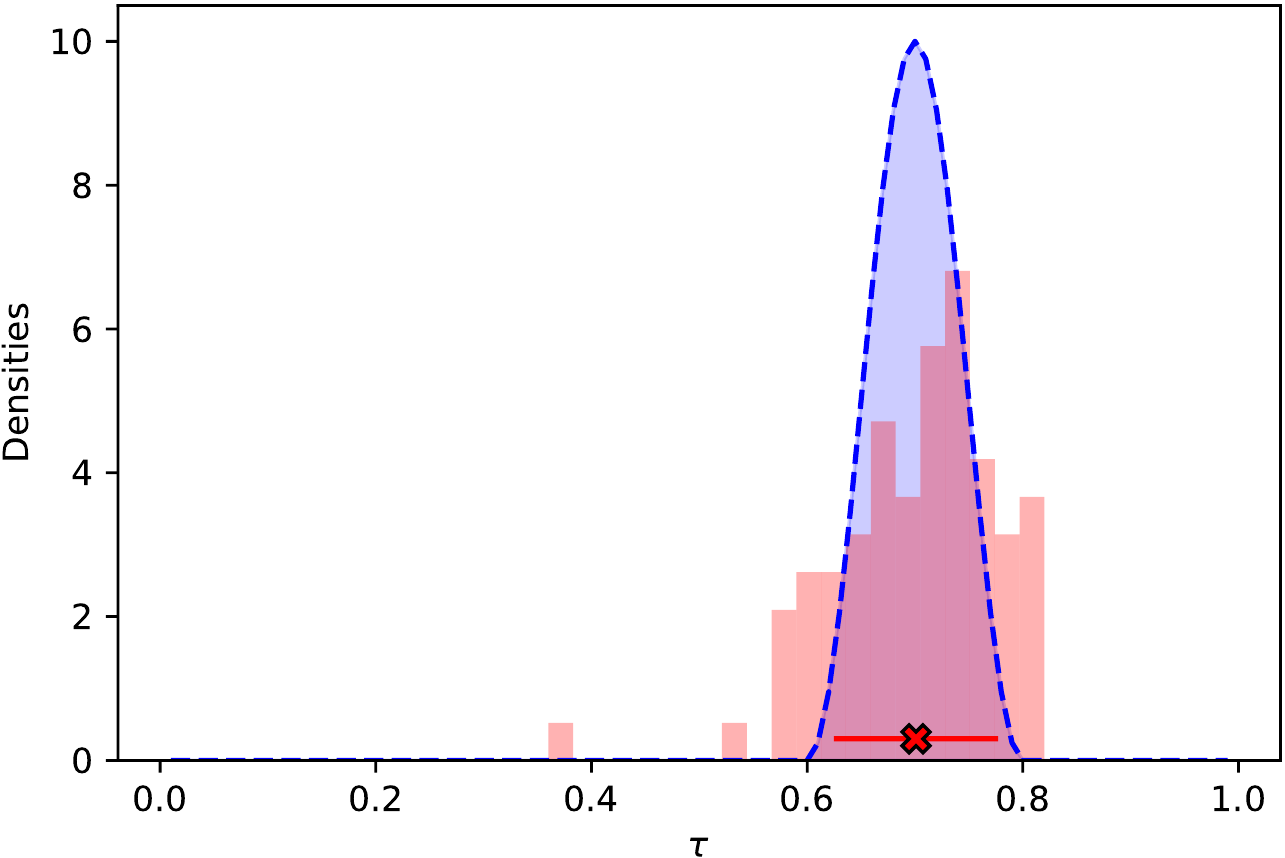} 
    \caption{Raised cosine: $\mu=0.7$, $\delta=0.1$.}
  \end{subfigure}
    \begin{subfigure}[t]{.4\linewidth}
    \centering
    \includegraphics[width=\linewidth]{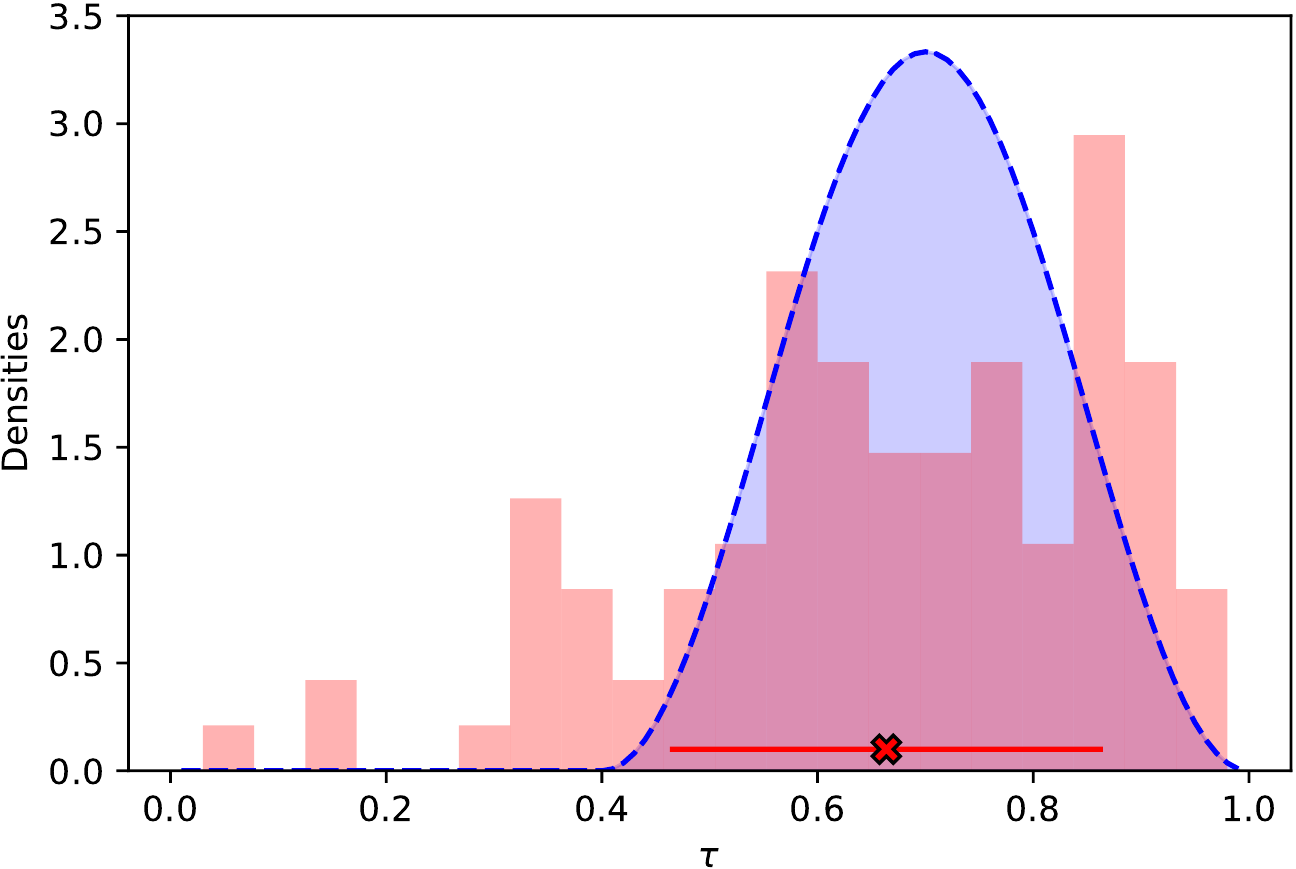}
    \caption{Raised cosine: $\mu=0.7$, $\delta=0.3$.}
  \end{subfigure}
  \end{array}
  $$
\caption{Comparison between theoretical and empirical distribution functions for the threshold. Dashed blue lines: uniform distribution and raised cosine distribution for different values of $\mu$ and $\delta$. Red histograms: the optimal thresholds provided by $100$ runs of the network when the loss is driven by $f_1$-score. Red crosses and red horizontal bars: average values of the histograms with standard deviations.}
\label{fig:adult}
\end{figure}

Let $\eta\in\mathbb{N}$, $\eta\le100$, be the number of successful training processes. Figure \ref{fig:adult} compares the probability distribution chosen \textit{a priori} for the threshold as in the left panel of Figure \ref{fig:distributions} (dashed blue line) to the \textit{empirical} probability distribution $f_{\tau^{\star}}$ (red histograms) derived from the optimal thresholds $\tau^{\star}_1,\dots,\tau^{\star}_{\eta}$ over the $\eta$ successful runs.  We modelled this distribution via normalized histograms. Moreover, we also displayed the average optimal threshold (red cross) resulting from the $\eta$ runs. The horizontal solid red line indicates the standard deviation related to the optimal thresholds. These experiments show that the optimal threshold is indeed heavily influenced by the chosen distribution, which effectively drives the optimization process to prefer a certain interval of threshold values.

\begin{table}[h]
\centering
\resizebox{0.99\textwidth}{!}{
\begin{tabular}{|c | c | c c c c c c |} 
\hline
    & \multirow{3}{*}{Uniform} & \multicolumn{6}{c||}{Raised cosine}\\ 
    \cline{3-8}
   & & \multicolumn{2}{c|}{$\mu=0.3$} & \multicolumn{2}{c|}{$\mu=0.5$} &  \multicolumn{2}{c||}{$\mu=0.7$} \\
   & & $\delta=0.1$ & \multicolumn{1}{c|}{$\delta=0.3$} & $\delta=0.1$ & \multicolumn{1}{c|}{$\delta=0.3$}  & $\delta=0.1$ & \multicolumn{1}{c||}{$\delta=0.3$}  \\
 \hline
\#success & $100$  & $90$ & \multicolumn{1}{c|}{$100$} & $100$ & \multicolumn{1}{c|}{$100$} & $83$ & $100$\\
\#out-of-support &  $0$ & $8$ & \multicolumn{1}{c|}{$13$} & $14$ & \multicolumn{1}{c|}{$9$} & $13$ & $13$ \\
\#epochs &  $119.46_{(\pm 43.95 )}$ & $125.21_{(\pm 44.39 )}$ & \multicolumn{1}{c|}{$121.05_{(\pm 46.14 )}$} & $124.07_{(\pm 42.27 )}$ & \multicolumn{1}{c|}{$131.12_{(\pm 48.89 )}$} & $124.00_{(\pm 44.97 )}$ & $123.08_{(\pm 38.51 )}$\\
$\tau^{\star}$ &  $0.53_{(\pm 0.36 )}$ & $0.31_{(\pm 0.06 )}$ & \multicolumn{1}{c|}{$0.35_{(\pm 0.22 )}$} & $0.50_{(\pm 0.07 )}$ & \multicolumn{1}{c|}{$0.52_{(\pm 0.19 )}$} & $0.70_{(\pm 0.08 )}$ & $0.66_{(\pm 0.20 )}$\\
f$_1$-score($\tau^{\star})$ &  $0.7510_{(\pm 0.01 )}$ & $0.7578_{(\pm 0.01 )}$ & \multicolumn{1}{c|}{$0.7556_{(\pm 0.01 )}$} & $0.7557_{(\pm 0.01 )}$ & \multicolumn{1}{c|}{$0.7571_{(\pm 0.01 )}$} & $0.7549_{(\pm 0.01 )}$ & $0.7546_{(\pm 0.01 )}$\\
 \hline
\end{tabular}
}
\caption{Outcomes of the application of the SOL functions in the case of a neural network trained on the {\em{Adult}} database. The chosen SOL relies on the $f_1$-score and the threshold $\tau$ is drawn according to the uniform and raised cosine distributions.}
\label{tab:prima}
\end{table}

\subsection{SOLs versus CE: addressing strongly imbalanced classes}
We then considered the problem of predicting high levels of pollution by using the Beijing air quality data set from the University of California at Irvine (UCI), available at $https://archive.ics.uci.edu/ml/datasets/Beijing+PM2.5+Data$.
The data set contains PM2.5 pollution
data and meteorological data including dew point, temperature, pressure, wind direction, wind speed and the cumulative number of hours of snow and rain. Data are collected every hour for five years, sourced from the data interface released by the US Embassy in Beijing \cite{Liang2015}.

We first fixed a level of PM2.5 concentration equal to $400$ and we labelled the feature vector containing the meteorological data and the PM2.5 concentration at time $T$ with class $1$ if the PM2.5 concentration at time $T+1$ is over the fixed level.
The data set ranges from 01/01/2010 to 12/31/2014: we considered $13104$ samples in training (about one year and a half) and $4320$ samples in testing (about half a year) by taking the testing period right after the training period. In order to guarantee statistical robustness we repeated $100$ times the construction of training and test sets, by shifting the training period and test period of $5$ days each time. The data set is strongly imbalanced since the mean rate of positive samples in the training set is about $1.41\%$ and in the testing set is about $1.19\%$. During the training phase, $1/3$ of the training set was used as validation set.

We used a fully-connected feed-forward NN with an input layer ($15$ ReLu neurons), a hidden layer ($8$ ReLu neurons) and a sigmoid output unit. We trained on $1000$ epochs and we imposed an early stopping condition when no improvement in the validation loss was obtained after $50$ epochs. As for loss functions, we considered the SOL functions related to the \textit{True Skill Statistic} (TSS) \cite{allouche2006assessing} and \textit{Critical Success Index} (CSI) \cite{schaefer1990critical}, 
which are very common skill scores adopted in meteorology applications and particularly appropriate in the case of strongly imbalanced data set. They are defined as
\begin{equation*}
    \textrm{TSS}=\frac{\textrm{TP}}{\textrm{TP}+\textrm{FN}}+\frac{\textrm{TN}}{\textrm{TN}+\textrm{FP}}-1,\quad
    \textrm{CSI}=\frac{\textrm{TP}}{\textrm{TP}+\textrm{FP}+\textrm{FN}}.
\end{equation*}
We point out that the TSS score satisfies the linearity assumption in Theorem \ref{thm:linea}, while the CSI does not.

Considering both the uniform and raised Cosine distributions $C(\mu,\delta)$ (in the case $\mu=0.5$ and $\delta=0.1,0.3$), we compared the proposed SOLs against the CE loss. For each run we performed the \textit{a posteriori} score maximization on the training set in order to compute the optimum threshold. In Table \ref{tab:poll_tss}  we reported the performances on both training and test sets by evaluating the skill scores with both the usual threshold $0.5$ and the optimum threshold $\tau^{\star}$ selected by the \textit{a posteriori} maximization. As in the previous application, we did not include cases in which the training optimization process was unsuccessful, and we reported the number of success on $100$ runs.

Table \ref{tab:poll_tss} shows that using the SOL related to a given skill score leads to better results with respect to the ones obtained by using the CE loss, in the case of both the training and the test sets. In particular, the skill scores are largely satisfactory by setting $\tau=0.5$ without the need to perform the a posteriori score maximization. In the case where the raised cosine distribution is implemented, this is due to the capability of the SOL function of driving the optimization process to promote thresholds in a neighborhood of the parameter $\mu$ (equal to $0.5$ in these experiments). In the case of the uniform distribution, no particular values of the threshold are promoted in the optimization process and this leads to results with high skill scores independently from the choice of the threshold.

Finally, we notice that the mean number of epochs to reach the convergence condition is lower for SOL functions with respect to the CE loss: in particular such a number is much lower for the SOL function in which the raised cosine distribution is chosen with $\delta=0.1$. 

\begin{table}[htb!]
\centering
\resizebox{0.99\textwidth}{!}{
\begin{tabular}{|c | c   c  c  c  c | c c|} 
\hline
& \multicolumn{5}{c|}{Training} & \multicolumn{2}{c|}{Test} \\
\cline{2-8}
Loss & \#success &
  \#epoch &
TSS & 
  $\tau^{\star}$ &
  TSS($\tau^{\star}$) & TSS & TSS($\tau^{\star}$) \\ 
   \hline
   CE &  100  &  $1000.00_{(\pm 0.00 )}$  &  $0.6130_{(\pm 0.26 )}$  &  $0.06_{(\pm 0.03 )}$  &  $0.9302_{(\pm 0.01 )}$  &  $0.5180_{(\pm 0.31 )}$ &  $0.8608_{(\pm 0.19 )}$ \\
TSS SOL ($\mathcal{U}$) &  100  &  $898.11_{(\pm 209.45 )}$  &  $0.9515_{(\pm 0.02 )}$  &  $0.65_{(\pm 0.15 )}$  &  $0.9579_{(\pm 0.01 )}$  &  $0.8634_{(\pm 0.20 )}$ &  $0.8681_{(\pm 0.19 )}$ \\
TSS SOL ($\mathcal{C}(0.5,0.1)$) &  87  &  $499.29_{(\pm 327.05 )}$  &  $0.9517_{(\pm 0.02 )}$  &  $0.52_{(\pm 0.02 )}$  &  $0.9568_{(\pm 0.02 )}$  &  $0.8699_{(\pm 0.18 )}$ &  $0.8649_{(\pm 0.18 )}$ \\
TSS SOL ($\mathcal{C}(0.5,0.3)$) &  94  &  $794.95_{(\pm 292.17 )}$  &  $0.9561_{(\pm 0.01 )}$  &  $0.57_{(\pm 0.06 )}$  &  $0.9624_{(\pm 0.01 )}$  &  $0.8641_{(\pm 0.20 )}$ &  $0.8617_{(\pm 0.20 )}$ \\
TSS SOL ($\mathcal{C}(0.5,0.5)$) &  99  &  $877.49_{(\pm 231.40 )}$  &  $0.9562_{(\pm 0.01 )}$  &  $0.61_{(\pm 0.08 )}$  &  $0.9616_{(\pm 0.01 )}$  &  $0.8676_{(\pm 0.19 )}$ &  $0.8653_{(\pm 0.19 )}$ \\
 \hline
\end{tabular}
}\vspace{0.5cm}
\resizebox{0.99\textwidth}{!}{
\begin{tabular}{|c | c   c  c  c  c | c c|} 
\hline
& \multicolumn{5}{c|}{Training} & \multicolumn{2}{c|}{Test} \\
\cline{2-8}
Loss & \#success &
  \#epoch &
CSI & 
  $\tau^{\star}$ &
  CSI($\tau^{\star}$) & CSI & CSI($\tau^{\star}$) \\ 
   \hline
   CE &  100  &  $1000.00_{(\pm 0.00 )}$  &  $0.5222_{(\pm 0.22 )}$  &  $0.44_{(\pm 0.09 )}$  &  $0.6201_{(\pm 0.06 )}$  &  $0.3583_{(\pm 0.23 )}$ &  $0.4345_{(\pm 0.18 )}$ \\
CSI SOL ($\mathcal{U}$) &  100  &  $697.03_{(\pm 258.69 )}$  &  $0.6718_{(\pm 0.12 )}$  &  $0.62_{(\pm 0.14 )}$  &  $0.6962_{(\pm 0.07 )}$  &  $0.4514_{(\pm 0.17 )}$ &  $0.4613_{(\pm 0.17 )}$ \\
CSI SOL ($\mathcal{C}(0.5,0.1)$) &  91  &  $398.26_{(\pm 221.41 )}$  &  $0.6769_{(\pm 0.12 )}$  &  $0.51_{(\pm 0.02 )}$  &  $0.6913_{(\pm 0.09 )}$  &  $0.4454_{(\pm 0.19 )}$ &  $0.4427_{(\pm 0.21 )}$ \\
CSI SOL ($\mathcal{C}(0.5,0.3)$) &  100  &  $528.04_{(\pm 251.06 )}$  &  $0.6720_{(\pm 0.12 )}$  &  $0.53_{(\pm 0.05 )}$  &  $0.6951_{(\pm 0.06 )}$  &  $0.4486_{(\pm 0.18 )}$ &  $0.4582_{(\pm 0.18 )}$ \\
CSI SOL ($\mathcal{C}(0.5,0.5)$) &  100  &  $618.63_{(\pm 267.47 )}$  &  $0.6698_{(\pm 0.14 )}$  &  $0.55_{(\pm 0.08 )}$  &  $0.6955_{(\pm 0.08 )}$  &  $0.4530_{(\pm 0.19 )}$ &  $0.4582_{(\pm 0.19 )}$ \\
 \hline
\end{tabular}
}
\caption{Comparison of binary crossentropy and SOL functions in the case of a neural network trained on the pollution dataset from UCI. The chosen SOL relies on the TSS (top table) and CSI (bottom table) scores and the threshold $\tau$ is drawn according to the uniform and raised cosine distributions. In columns "TSS" and "CSI" the scores are computed with respect to the threshold $\tau=0.5$.
}
\label{tab:poll_tss}
\end{table}

\section{Conclusions}\label{sec:conclusions}

In this paper, we discussed a new approach for the automatic maximization of a chosen score in classification tasks. After providing a detailed theoretical construction, we showed the effectiveness as well as the consistency of our framework. In particular, we described how this approach naturally allows the introduction of loss functions whose explicit form depends on the form of the skill scores defined on probabilistic confusion matrices. The performances of these loss functions during the training phase of supervised networks are tested in the case of two experimental forecasting problems. 

We point out that the SOL functions that we have here introduced in the binary classification setting can be extended to the multi-class case, i.e. for $d>2$, in a rather straightforward fashion. In fact, the idea is to use a standard one-versus-rest approach in which $d$ different $2 \times 2$ matrices $\mathrm{CM}_1(\tau_1,\mathcal{S}(\bs{w})),\dots,\mathrm{CM}_d(\tau_d,\mathcal{S}(\bs{w}))$ are derived in the case of $\tau_1,\dots,\tau_d$ different random variables. In this way, the investigation of the multi-class case reduces to one of the binary setting.

As far as future work is concerned, we are currently following two paths. 

From a methodological viewpoint, we think that the SOL functions approach should be extended to the case of value-weighted skill scores that allow the assessment of prediction errors on the basis of their impact on the decision making process \cite{guastavino2021}. 

From the viewpoint of applications, our main interest is to exploit this approach in the framework of space weather \cite{camporeale2018machine} and, in particular, for the prediction of extreme solar flaring storms, whereby the training sets of solar magnetograms used for training the networks are strongly imbalanced against highly energetic events \cite{benvenuto2020machine}.

\section{Acknowledgements}

SG is financially supported by a regional grant of the `Fondo Sociale Europeo', Regione Liguria. FM is supported by the financial contribution from the agreement ASI-INAF n.2018-16-HH.0. MP and CC acknowledges this same agreement. This research has been partly accomplished within the Rete ITaliana di Approssimazione (RITA) and with the support of GNCS-IN$\delta$AM.

\bibliographystyle{siamplain}
\bibliography{references_papero}
\end{document}